\documentclass[conference]{IEEEtran}
\usepackage{times}

\usepackage[numbers]{natbib}
\usepackage{multicol}
\usepackage[bookmarks=true]{hyperref}

\usepackage{amsmath}
\usepackage{amsfonts}

\usepackage{graphicx}

\usepackage{threeparttable}
\usepackage{multirow}
\usepackage{booktabs}

\usepackage{makecell}
\usepackage{arydshln}
\usepackage{mathtools}

\begin{document}

\title{
Traj-LIO: A Resilient Multi-LiDAR Multi-IMU State Estimator Through Sparse Gaussian Process
}

\author{\authorblockN{Xin~Zheng}
\authorblockA{College of Computer Science\\ Zhejiang University\\ Hangzhou, China, 310027\\
Email: xinzheng@zju.edu.cn}
\and
\authorblockN{Jianke~Zhu}
\authorblockA{College of Computer Science\\ Zhejiang University\\ Hangzhou, China, 310027\\
Email: jkzhu@zju.edu.cn}}

\maketitle

\begin{abstract}
Nowadays, sensor suits have been equipped with redundant LiDARs and IMUs to mitigate the risks associated with sensor failure. It is challenging for the previous discrete-time and IMU-driven kinematic systems to incorporate multiple asynchronized sensors, which are susceptible to abnormal IMU data. To address these limitations, we introduce a multi-LiDAR multi-IMU state estimator by taking advantage of Gaussian Process (GP) that predicts a non-parametric continuous-time trajectory to capture sensors’ spatial-temporal movement with limited control states. Since the kinematic model driven by three types of linear time-invariant stochastic differential equations are independent of external sensor measurements, our proposed approach is capable of handling different sensor configurations and resilient to sensor failures. Moreover, we replace the conventional $\mathrm{SE}(3)$ state representation with the combination of  $\mathrm{SO}(3)$ and vector space, which enables GP-based LiDAR-inertial system to fulfill the real-time requirement. Extensive experiments on the public datasets demonstrate the versatility and resilience of our proposed multi-LiDAR multi-IMU state estimator. To
contribute to the community, we will make our source code publicly available.

\end{abstract}

\IEEEpeerreviewmaketitle

\section{Introduction}
State estimation~\cite{barfoot2024state} is a fundamental task in robotics, which predicts the underlying state of system through a sequence of measurements from various sensors. Over the past decades, LiDAR has emerged as one of the most widely adopted exteroceptive sensors for pose estimation due to its excellent range sensing capability in unstructured scenarios~\cite{thrun2006stanley} and low-illumination environments~\cite{ebadi2022present}. In the practical studies~\cite{helmberger2022hilti,zhang2022hilti,nguyen2022ntu}, the inertial-aided odometry systems~\cite{reinke2022locus,shan2020lio,xu2022fast,chen2023direct} are considered as more reliable state estimator by coupling the interoceptive Inertial Measurement Units (IMUs).  

Generally, a single LiDAR has a limited vertical Field of View (FoV), leading to inadequate geometric constraints in the degenerated cases~\cite{tuna2023x}. Low-cost IMUs are sensitive to temperature variations and mechanical shocks, which may yield inaccurate measurements. Consequently, current autonomous robots~\cite{reinke2022locus,nguyen2022ntu} are equipped with redundant multi-LiDAR and multi-IMU sensor suits to mitigate the risks of inevitable sensor failure. However, existing LiDAR-inertial Odometry (LIO)s~\cite{zhang2014loam,shan2020lio,xu2022fast,chen2023direct} are built upon discrete-time and IMU-driven machinery, which struggle to fuse different kinds of asynchronized sensors.

The discrete-time estimator requires to add new state variables at each measurement time, which leads to unbounded optimization complexity, especially in the case of several asynchronized sensors at different frequencies in the system. For LiDAR sensors, continuously captured points between two consecutive discrete states have to be compensated for motion distortions~\cite{zhang2014loam,xu2022fast} by assuming to be measured simultaneously. Thus, the accuracy of LiDAR odometry heavily depends on IMU-driven kinematics~\cite{sola2017quaternion,forster2015manifold}, which are employed to not only undistort the points but also provide an initial propagated state for registration. Therefore, several solutions~\cite{bosse2012zebedee,nguyen2023slict,lv2023continuous,ramezani2022wildcat,lv2021clins} have replaced the discrete-time states with the parametric continuous-time trajectory~\cite{furgale2012continuous}, enabling the querying of specific states given a timestamp. Unfortunately, these LIOs still rely on IMU information for point cloud preprocessing~\cite{bosse2012zebedee,ramezani2022wildcat} or providing motion constrains~\cite{nguyen2023slict,lv2021clins,lv2023continuous}. IMU failure can directly lead to system crashes. While switching to another valid IMU upon detecting their status is feasible, current IMU-driven systems, whether based on the Kalman Filter~\cite{sola2017quaternion} or pre-integration~\cite{forster2015manifold}, lack an effective method to fully utilize multi-IMU information in state estimation.

\begin{figure}[t]
	\centering
\includegraphics[width=0.48\textwidth]{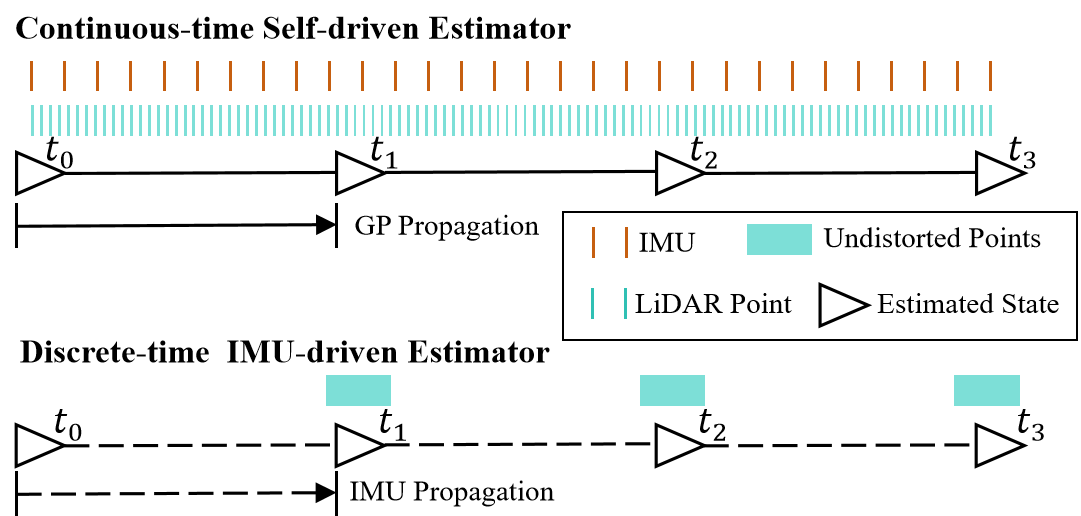}
	\caption{The schematic of continuous-time self-driven estimator vs discrete-time IMU-driven scheme. In the discrete-time estimator, continuously captured LiDAR points must be transferred to a specific state, while our continuous-time estimator can query any state through GP interpolation. Besides, the IMU-driven estimator relies on IMU data for state propagation, while the kinematics of our estimator is driven by GP prior.}
	\label{fig:gp_estimator}
	\vspace{-0.2in}
\end{figure}

An ideal LiDAR-inertial state estimator should be versatile across different sensor configurations and resilient to inevitable sensor failures. To overcome the aforementioned limitations, we propose a multi-LiDAR multi-IMU state estimator through sparse Gaussian Process (GP)~\cite{williams2006gaussian,barfoot2014batch}, as shown in Fig.~\ref{fig:gp_estimator}. Firstly, GP provides a non-parametric representation of continuous-time trajectory, effectively incorporating multiple asynchronized sensors. With the limited control states, it is capable of querying any interested states on the trajectory through GP interpolation~\cite{barfoot2014batch,anderson2015full,dong2018sparse}. Secondly, a particular class of GP kernels~\cite{anderson2015full,tang2019white,wong2020data} enables us to derive a self-driven kinematic system that does not rely on either an IMU input or other specific sensors. All input information from different sensors are treated as system measurements, which makes our proposed estimator resilient to sensor failure. Unlike previous GP method~\cite{wu2022picking} that uses computationally demanding $\mathrm{SE}(3)$ state representations, we decouple the state space into rotation in $\mathrm{SO}(3)$ and translation in vector space. This approach simplifies the derivation of analytic Jacobians and avoids complexities associated with $\mathrm{SE}(3)$. Moreover, our presented state representation facilitates direct integration with IMU measurements. Although this paper focuses on multi-LiDAR multi-IMU setting, the proposed estimator can be seamlessly extended to other kinds of sensor, such as camera, GPS, wheel encoder and etc.

To evaluate the effectiveness and generalizability of our proposed LiDAR-inertial state estimator, we conduct experiments on public datasets from handheld scenario~\cite{zhang2022hilti} to aggressive UAV situations~\cite{nguyen2022ntu}, even in scenarios where the kinematic state exceeds the IMU’s measurement range~\cite{he2023point}. Our self-driven kinematic system demonstrates resilience in various sensor failures and achieves competitive accuracy. To contribute to the community, we will make our implementation publicly available, allowing to easily integrate other types of sensors into our proposed framework.

In summary, the main contributions of this paper are: 1) a versatile continuous-time multi-LiDAR multi-IMU state estimator through Gaussian Process, which is adaptable to various kinds of sensor configuration; 2) a resilient kinematic model based on GP prior, which is self-driven and independent of IMU; 3) a real-time GP-based LiDAR-inertial odometry method with analytic Jacobians through decoupling the state space into rotation in $\mathrm{SO}(3)$ and translation in vector space.

\section{Related Works}
In this section, we review the related literature on LiDAR odometry systems that can be roughly categorized into two groups, including discrete-time LiDAR-inertial methods and continuous-time approaches.

\subsection{Discrete-time LiDAR-Inertial System}
The discrete-time LiDAR-Inertial system represents the continuous movement with a series of discrete states. \citet{zhang2014loam} propose the seminal work LOAM that estimates poses by a modified Iterative Closest Point (ICP) algorithm~\cite{besl1992method} incorporating both point-to-line and point-to-plane metrics. To overcome the motion distortion between two consecutive states, the additional IMU is used for motion compensation and provides an initial guess for point registration. Since the LiDAR and IMU are not in a joint optimization scheme, these approaches~\cite{zhang2014loam,chen2022direct,chen2023direct,palieri2020locus,reinke2022locus} are regarded as loosely-coupled LIO. \citet{chen2022direct,chen2023direct} replace the above feature-based ICP by Generalized ICP~\cite{segal2009generalized} so that it can be deployed on resource-limited platforms. LOCUS~\cite{palieri2020locus,reinke2022locus} is a multi-sensor framework that merges multi-LiDAR into single point cloud for scan matching.

In contrast to loosely-coupled methods, tightly-coupled solutions jointly optimize the sensor measurement to improve the odometry accuracy. \citet{ye2019tightly} learn from the successful visual-inertial system~\cite{qin2018vins} and jointly optimize the LiDAR and pre-integrated IMU measurements. \citet{li2021towards} extend it to non-repetitive LiDAR and suggest a new feature selection procedure. \citet{shan2020lio} formulate LIO by a factor graph framework to integrate various constraints, including loop closure, wheel encoder and GPS, into the LIO system. Comparing to these multi-state estimators, filter-based LIOs are more efficient. \citet{qin2020lins} firstly fuse LiDAR with IMU measurements through an iterative error-state Kalman Filter framework~\cite{sola2017quaternion}. \citet{xu2022fast} suggest an efficient Kalman gain and directly register raw points into an incremental kd-tree map to facilitate fast neighborhood searching. \citet{jung2023asynchronous} extend it to fuse multiple asynchronous LiDARs. All of the aforementioned LIOs rely on IMU-driven kinematics~\cite{sola2017quaternion}, which easily crashes with abnormal IMU data. \citet{he2023point} augment acceleration and angular velocity into state space and employ a pointwise registration scheme, improving the system's robustness in extreme scenarios. 

Since the discrete-time state estimators struggle to deal with multiple asynchronous sensors, their optimization complexity increases dramatically along with the number of sensors. 

\subsection{Continuous-time LiDAR-Inertial System}
In contrast to discrete-time method, the continuous-time trajectory~\cite{furgale2012continuous} provides a more elegant way to represent sensors' spatial-temporal movement. Given the timestamp, the interested state can be queried immediately. The challenge is how to approximate the entire trajectory with finite states while ensuring both accuracy and efficiency. 

One major class of continuous-time LIOs makes use of the parametric trajectory. Zebedee~\cite{bosse2012zebedee}, ElasticSLAM~\cite{park2021elasticity}, and Wildcat~\cite{ramezani2022wildcat} adopt a piecewise linear function to represent trajectories, which offer a straightforward to query state by linear interpolation. To eliminate the error due to linear assumption, they require a series of high-frequency control poses in aggressive motion. \citet{nguyen2023slict} fuse this continuous-time trajectory representation with pre-integration~\cite{forster2015manifold} to develop a multi-LiDAR and single IMU odometry. Moreover, \citet{lv2021clins} introduce B-spline into tightly coupled LIO. However, it has difficulties in fulfilling real-time requirements even with more efficient parameterization~\cite{sommer2020efficient}. Its successor, CLIC~\cite{lv2023continuous}, extends the B-spline representation into the LiDAR-inertial-camera system, which can fuse multi-LiDAR with single IMU.

Another class of continuous-time representation is based on non-parametric Gaussian Processes~\cite{williams2006gaussian}. Tong et al.~\cite{tong2013gaussian} employ a Gaussian Process Gauss-Newton optimization to estimate the 2D continuous movement.~\citet{barfoot2014batch} find the sparsity of the inverse kernel matrix generated by the linear time-varying stochastic differential equations, which make GP regression more efficient.~\citet{anderson2015full} bring the batch GP continuous-time trajectory estimation into $\mathrm{SE}(3)$ space, while~\citet{dong2018sparse} extend it to the general Lie group. However, current GP-based state estimator is limited to LiDAR-only odometry~\cite{wu2022picking}, which has difficulty in achieving the real-time performance, not to mention multi-LiDAR and multi-IMU fusion. In this work, we suggest an effective GP-based LiDAR-inertial odometry. States of rotation, translation and IMU bias are separately modeled by three types of GP motion priors, which achieves the minimal state representation and avoids complexities associated with $\mathrm{SE}(3)$ in the derivation of analytic Jacobians.

\section{Sparse Gaussian Process Revisited}~\label{sec:gp}
In contrast to conventional discrete-time optimization solutions~\cite{shan2020lio, xu2022fast}, continuous-time state estimator~\cite{furgale2012continuous} efficiently integrates multiple synchronized sensors by querying states at the specific timestamps. Instead of using an explicitly parameterized state trajectory~\cite{bosse2012zebedee, lv2021clins, ramezani2022wildcat, zheng2023traj}, we employ a flexible non-parametric approach to represent the continuous-time state as a one-dimensional Gaussian Process~\cite{williams2006gaussian} by treating time as the independent variable. Specifically, we model the continuous-time state $\mathbf{x}(t)$ over a temporal window $[t_{0}, t_{K})$ as follows:
\begin{align}~\label{equ:ct-state}
\mathbf{x}(t) \sim  \mathcal{GP}(\boldsymbol{\mu}(t), \boldsymbol{\mathcal{K}}(t, t^{'}) ), \quad t, t^{'} \in [t_{0}, t_{K}),
\end{align}
where $\boldsymbol{\mu}(t)$ denotes the mean function, and $\boldsymbol{\mathcal{K}}(t, t^{'})$ represents the covariance function. We initialize the mean state and prior covariance matrix as $\boldsymbol{\mu}(t_{0}) = \boldsymbol{\mu}_{0}$ and $\boldsymbol{\mathcal{K}}(t_{0}, t_{0}) = \boldsymbol{\mathcal{K}}_{0}$, respectively.

In this paper, the continuous-time state $\mathbf{x}(t)$ extends beyond the mere poses (translation and rotation)~\cite{barfoot2014batch, anderson2015full, dong2018sparse} to velocity, acceleration, and IMU bias. By governing each kind of state through a specific GP prior, the state estimation can be formulated as a GP regression problem~\cite{tong2013gaussian}. This section revisits three GP priors used in our multi-LiDAR multi-IMU state estimator as well as the related GP interpolation.

\subsection{Gaussian Process Prior on Vector Space}~\label{sec:vector}

\begin{table*}[t]
\renewcommand{\arraystretch}{1.3} 
\centering
\caption{LTI system $\dot{\mathbf{x}} (t)=\mathbf{A}(t)\mathbf{x}(t)+\mathbf{F}(t)\mathbf{w}(t)$ based on three motion priors, where $\Delta t=t-t_{0}$}
\label{tab:prior}
\begin{threeparttable}
\begin{tabular}{@{}ccccccc@{}}
\toprule
\midrule
GP Prior& SDE & $\mathbf{x}(t)$&$\mathbf{A}(t)$ &  $\mathbf{F}(t)$& $\boldsymbol{\Phi}(t,t_{0})$ & $\mathbf{Q}(t,t_{0})$\\
\midrule
Random Walk &$\dot{\mathbf{p}}(t) =\mathbf{w}(t)$ & $\mathbf{p}(t)$ & $\mathbf{0}$& $\mathbf{I}$& $\mathbf{I}$&$\Delta t\mathbf{Q}_{c}$\\
[5pt]
Constant Velocity &$ \ddot{\mathbf{p}}(t) =\mathbf{w}(t)$ & $\begin{bmatrix}
 \mathbf{p}(t) \\
\dot{ \mathbf{p}}(t)
\end{bmatrix}$ & $\begin{bmatrix}
  \mathbf{0} &\mathbf{I} \\
   \mathbf{0}& \mathbf{0}\\
\end{bmatrix}$& $\begin{bmatrix}
 \mathbf{0}\\
\mathbf{I}
\end{bmatrix}$& $\begin{bmatrix}
  \mathbf{I} & \Delta t\mathbf{I}  \\
  \mathbf{0}&  \mathbf{I}\\
\end{bmatrix}$&$\begin{bmatrix}
  \frac{1}{3}\Delta t^{3}\mathbf{Q}_{c} & \frac{1}{2}\Delta t^{2}\mathbf{Q}_{c}\\
  \frac{1}{3}\Delta t^{3}\mathbf{Q}_{c}&\Delta t\mathbf{Q}_{c}
\end{bmatrix}$\\
[10pt]
Constant Acceleration &$ \dddot{\mathbf{p}}(t) =\mathbf{w}(t)$ & $\begin{bmatrix}
 \mathbf{p}(t) \\
 \dot{ \mathbf{p}}(t)\\
\ddot{ \mathbf{p}}(t)
\end{bmatrix}$ & $\begin{bmatrix}
  \mathbf{0} & \mathbf{I} &\mathbf{0} \\
  \mathbf{0}&  \mathbf{0}& \mathbf{I}\\
  \mathbf{0}&  \mathbf{0}&\mathbf{0}
\end{bmatrix}$& $\begin{bmatrix}
 \mathbf{0}\\
 \mathbf{0}\\
\mathbf{I}
\end{bmatrix}$& $\begin{bmatrix}
  \mathbf{I} & \Delta t\mathbf{I} &\frac{1}{2}\Delta t^{2} \mathbf{I} \\
  \mathbf{0}&  \mathbf{I}& \Delta t\mathbf{I}\\
  \mathbf{0}&  \mathbf{0}&\mathbf{I}
\end{bmatrix}$&$\begin{bmatrix}
  \frac{1}{20}\Delta t^{5}\mathbf{Q}_{c} & \frac{1}{8}\Delta t^{4}\mathbf{Q}_{c}& \frac{1}{6}\Delta t^{3}\mathbf{Q}_{c}\\
  \frac{1}{8}\Delta t^{4}\mathbf{Q}_{c} & \frac{1}{3}\Delta t^{3}\mathbf{Q}_{c}& \frac{1}{2}\Delta t^{2}\mathbf{Q}_{c} \\
  \frac{1}{6}\Delta t^{3}\mathbf{Q}_{c} & \frac{1}{2}\Delta t^{2}\mathbf{Q}_{c}& \Delta t\mathbf{Q}_{c}
\end{bmatrix}$\\

\midrule   
\bottomrule
\end{tabular}
\end{threeparttable}
\end{table*}

\begin{table*}[t]
\renewcommand{\arraystretch}{1.5} 
\centering
\caption{Gaussian Interpolation $\mathbf{x}(\tau)=\boldsymbol{\Lambda}(\tau) \mathbf{x}(t_{k-1})+\boldsymbol{\Psi} (\tau)\mathbf{x}(t_{k})$, where $\Delta t=t_{k}-t_{k-1}, \alpha=(\tau-t_{k-1})/\Delta t$}
\label{tab:gp-inte}
\resizebox{\textwidth}{!}{
\begin{threeparttable}
\begin{tabular}{@{}c|c|c|c@{}}
\toprule
\midrule
& Random Walk & Constant Velocity&Constant Acceleration\\
\midrule
$\boldsymbol{\Lambda}(\tau)$ &$(1-\alpha)\mathbf{I}$& $\begin{bmatrix}
(1 + 2 \alpha)( 1-\alpha)^2 \mathbf{I} & \alpha(1-\alpha)^2  \Delta t\mathbf{I} \\
\frac{-6 \alpha(1-\alpha )}{\Delta t}\mathbf{I} & ( 1 - 3 \alpha)(1-\alpha )\mathbf{I}
\end{bmatrix}$&
$\begin{bmatrix}
 (1 + 3 \alpha + 6 \alpha^2)( 1-\alpha)^3\mathbf{I}  &  ( \alpha + 3 \alpha^{2})( 1-\alpha)^3 \Delta t\mathbf{I} & \frac{  \alpha^2( 1-\alpha)^3 \Delta t^2}{2} \mathbf{I}\\
\frac{-30 \alpha^2 (1-\alpha )^2}{\Delta t}\mathbf{I} &  (1 + 2 \alpha - 15 \alpha^2) (1 - \alpha)^2\mathbf{I}& \frac{ (2 \alpha - 5 \alpha^{2})(1 - \alpha)^2 \Delta t}{2} \mathbf{I} \\
\frac{-60 \alpha(1  - 2\alpha) (1-\alpha)}{\Delta t^2}\mathbf{I} & \frac{-12\alpha(3-5\alpha)(1-\alpha) }{\Delta t}\mathbf{I} & (1-8\alpha+10\alpha^2)(1-\alpha)\mathbf{I}
\end{bmatrix}$\\
\midrule
$\boldsymbol{\Psi}(\tau)$&$\alpha\mathbf{I}$&$\begin{bmatrix}
(3\alpha^2  - 2  \alpha^3)\mathbf{I} & (-\alpha^2 + \alpha^3)  \Delta t \mathbf{I}\\
\frac{6 \alpha( 1- \alpha )}{\Delta t}\mathbf{I} &  (-2\alpha + 3 \alpha^2)\mathbf{I}
\end{bmatrix}$&
$ \begin{bmatrix}
(10 - 15 \alpha + 6 \alpha^2)\alpha^3  \mathbf{I} & (-4 + 7 \alpha - 3 \alpha^2)\alpha^3  \Delta t \mathbf{I}& \frac{(1-\alpha )^2 \alpha^3 \Delta t^2}{2}\mathbf{I} \\
\frac{30 (1- \alpha)^2 \alpha^2}{\Delta t}\mathbf{I} &  (-12 + 28 \alpha - 15 \alpha^2)\alpha^2 \mathbf{I}& \frac{ (3 - 8 \alpha + 5 \alpha^2)\alpha^2 \Delta t}{2} \mathbf{I}\\
\frac{60 ( \alpha - 3 \alpha^2 + 2 \alpha^3)}{\Delta t^2}\mathbf{I} & \frac{-12 (2\alpha  - 7 \alpha^2 + 5 \alpha^3)}{\Delta t} \mathbf{I}& (3\alpha  - 12 \alpha^2 + 10 \alpha^3)\mathbf{I}
\end{bmatrix}$\\

\midrule   
\bottomrule
\end{tabular}
\end{threeparttable}
}
\end{table*}

\citet{barfoot2014batch} demonstrate that GPs generated by linear time-varying (LTV) stochastic differential equations (SDEs) possess an exactly sparse inverse kernel matrix. Such scheme is beneficial to GP regression and interpolation. Accordingly, we build our model using a stochastic LTV SDE without control input, as delineated by:
\begin{equation}~\label{equ:sde}
    \dot{\mathbf{x}} (t) = \mathbf{A}(t)\mathbf{x}(t) + \mathbf{F}(t)\mathbf{w}(t),
\end{equation}
where $\mathbf{A}(t)$ and $\mathbf{F}(t)$ represent the time-varying system matrices, and $\mathbf{w}(t)$ is the white process noise. This noise is characterized as a zero-mean white GP:
\begin{equation}
    \mathbf{w}(t) \sim \mathcal{GP}(\mathbf{0}, \mathbf{Q_{c}}\delta (t - t^{'})  ),  
\end{equation}
where $\mathbf{Q_{c}}$ is the power-spectral density matrix of the system, and $\delta (\cdot)$ denotes the Dirac delta function. Under this assumption, we can obtain the explicit form of the GP in Eq.~\ref{equ:ct-state}. The mean function is expressed as:
\begin{equation}
    \boldsymbol{\mu}(t) = \boldsymbol{\Phi}(t, t_{0}) \boldsymbol{\mu}_{0}, 
\end{equation}
where $ \boldsymbol{\Phi}(t, t_{0}) $ is the transition matrix of the LTV system. The covariance function at $t^{'} = t$ is:
\begin{equation}
    \begin{array}{c}
  \boldsymbol{\mathcal{K}} (t, t)  =  \boldsymbol{\Phi}(t, t_{0}) \boldsymbol{\mathcal{K}} _0  \boldsymbol{\Phi}(t, t_{0})^\top + \mathbf{Q}(t, t_{0}),  \\
\mathbf{Q}(t, t_{0})  =  \int_{t_0}^{t} \boldsymbol{\Phi}(t, s) \mathbf{F}(s) \mathbf{Q}_{c}  \mathbf{F}(s)^\top \boldsymbol{\Phi}(t, s) ^\top ds,
\end{array}
\end{equation} 
For values at $t^{'} \ne t$, its detailed derivations are presented in~\cite{barfoot2014batch}.

Practically, we can select several GP priors to simplify the above LTV system into a linear time-invariant (LTI) system. Our state estimator encompasses three GP priors, including the \textbf{Random Walk Process}, \textbf{Constant-Velocity Process}~\cite{barfoot2014batch}, and \textbf{Constant-Acceleration Process}~\cite{tang2019white}. The details of the LTI system with these three priors are listed in Table~\ref{tab:prior}. $\mathbf{p}(t)$ represents an N-dimensional state vector, which could denote either IMU bias or position. $\mathbf{x}(t)$ represents the augmented states that are Markovian.

\subsection{Gaussian Process Prior on Manifold}
Unfortunately, 3D rotation lies on an $\mathrm{SO}(3)$ manifold, which precludes the direct application of motion priors from vector spaces. The relationship between the rotation $\mathbf{R}(t)$ and its corresponding body-frame angular velocity $\boldsymbol{\omega}(t)$ is governed by a nonlinear SDE:
\begin{equation}~\label{equ:rot_sde}
    \dot{\mathbf{R}}(t) = \mathbf{R}(t)\boldsymbol{\omega}(t)^{\wedge},
\end{equation}
where the operator $\wedge$ represents the skew-symmetric matrix. To leverage our previous derivation for the LTI system, we adopt the technique in~\cite{anderson2015full,dong2018sparse} to linearize the global SDE within the local tangent space $\mathfrak{so}(3)$ of the $\mathrm{SO}(3)$ manifold. For a given global state $\mathbf{R}(t)$ over the interval $[t_{k-1}, t_{k})$, we define the local state variable $\boldsymbol{\theta}_{k}(t)$ around $\mathbf{R}_{k-1} = \mathbf{R}(t_{k-1})$ as follows:
\begin{equation}~\label{equ:local1}
    \boldsymbol{\theta}_{k}(t) = \mathrm{Log}(\mathbf{R}^{-1}_{k-1}\mathbf{R}(t)), \quad t \in [t_{k-1}, t_{k}),
\end{equation}
where $\mathrm{Log}: \mathrm{SO}(3) \rightarrow \mathbb{R}^{3}$ denotes the logarithmic mapping function~\cite{sola2018micro}. Substituting this into Eq.~\ref{equ:rot_sde}, we derive the local angular velocity $\dot{\boldsymbol{\theta}}_{k}(t)$ as:
\begin{equation}~\label{equ:local2}
    \dot{\boldsymbol{\theta}}_{k}(t) = \mathcal{J}_{r}(\boldsymbol{\theta}(t))^{-1}\boldsymbol{\omega}(t),
\end{equation}
$\mathcal{J}_{r}(\cdot)$ is the right Jacobian of $\mathrm{SO}(3)$~\cite{sola2018micro}. Consequently, the rotational state can be represented by a local LTI SDE:
\begin{equation}
    \dot{\boldsymbol{\gamma}}_{k}(t) = \mathbf{A}(t)\boldsymbol{\gamma}_{k}(t) + \mathbf{F}(t)\mathbf{w}(t).
\end{equation}~\label{equ:local_state}
where $\boldsymbol{\gamma}_{k}(t)$ denotes the local Markov state. If the motion prior is either a Random Walk Process $\dot{\boldsymbol{\theta}}_{k}(t) = \mathbf{w}(t)$ or a Constant Velocity Process $\ddot{\boldsymbol{\theta}}_{k}(t) = \mathbf{w}(t)$, the augmented state becomes:
\begin{equation}
   \boldsymbol{\gamma}_{k}(t) = \boldsymbol{\theta}_{k}(t), \quad \text{or} \quad \boldsymbol{\gamma}_{k}(t) = \begin{bmatrix}
\boldsymbol{\theta}_{k}(t) \\
\dot{\boldsymbol{\theta}}_{k}(t)
\end{bmatrix}.
\end{equation}
The system matrices $\mathbf{A}(t)$ and $\mathbf{F}(t)$, along with the transition and covariance matrices, are analogous to those in vector space, as listed in Table~\ref{tab:prior}. Given that the gyroscope measures angular velocity, the constant acceleration prior is unnecessary for the rotational state.

\subsection{Gaussian Process Interpolation}
The continuous-time trajectory allows to query the state at a given timestamp, which significantly reduces the complexity of optimization in handling data from multiple high-frequency sensors. GPs with previous motion priors exhibit a sparse structure to enhance the efficiency of interpolation~\cite{barfoot2014batch}. Considering a query time $\tau$ within the interval $[t_{k-1}, t_{k})$, we can deduce the state $\mathbf{x}(\tau)$ based on the initial state $\mathbf{x}(t_{k-1})$ and the final state $\mathbf{x}(t_{k})$ with O(1) complexity:
\begin{equation}
    \mathbf{x}(\tau) = \boldsymbol{\Lambda}(\tau) \mathbf{x}(t_{k-1}) + \boldsymbol{\Psi}(\tau)\mathbf{x}(t_{k}).
\end{equation}
The matrices $\boldsymbol{\Lambda}(\tau)$ and $\boldsymbol{\Psi}(\tau)$ are given by:
\begin{equation}
\begin{array}{c}
\boldsymbol{\Lambda}(\tau) =\boldsymbol{\Phi} (\tau,t_{k-1})-\boldsymbol{\Psi} (\tau)\boldsymbol{\Phi} (t_{k},t_{k-1}),\\
\boldsymbol{\Psi} (\tau)=\mathbf{Q}(\tau, t_{k-1}) \boldsymbol{\Phi} (t_{k},\tau)^{\top}\mathbf{Q}(t_{k},t_{k-1})^{-1}.
\end{array}
\end{equation}
Since we have provided transition matrix $\boldsymbol{\Phi}$ and covariance matrix $\mathbf{Q}$ in Table~\ref{tab:prior}, it is quite straightforward to calculate the exact forms of these coefficient matrices, as outlined in Table~\ref{tab:gp-inte}. Notably, the coefficient for the Random Walk process corresponds to the linear interpolation scheme utilized in parametric continuous-time LiDAR-(inertial) Odometry~\cite{bosse2012zebedee, dellenbach2022ct, zheng2023traj, ramezani2022wildcat}. For the Constant Velocity Process, it aligns with cubic Hermite polynomial interpolation~\cite{barfoot2024state}.

Based on the aforementioned formulation, we can directly interpolate states in vector space, such as position, velocity, acceleration, and IMU bias. For rotation $\mathbf{R}(t)$ and angular velocity $\boldsymbol{\omega}(t)$, it is necessary firstly to convert them into local state variables as below:
$$
\boldsymbol{\gamma}_{k}(t_{k-1}) = \begin{bmatrix}
\mathbf{0} \\
\boldsymbol{\omega}(t_{k-1})
\end{bmatrix},
\boldsymbol{\gamma}_{k}(t_{k}) = \begin{bmatrix}
\boldsymbol{\theta}_{k}(t_{k}) \\
\mathcal{J}_{r}(\boldsymbol{\theta}_{k}(t_{k}))^{-1}\boldsymbol{\omega}(t_{k})
\end{bmatrix}.
$$
Subsequently, GP interpolation on the local state is:
\begin{equation}
\boldsymbol{\gamma}_{k}(\tau) = \begin{bmatrix}
\boldsymbol{\theta}_{k}(\tau) \\
\dot{\boldsymbol{\theta}}_{k}(\tau)
\end{bmatrix} = \boldsymbol{\Lambda}(\tau) \boldsymbol{\gamma}_{k}(t_{k-1}) + \boldsymbol{\Psi}(\tau)\boldsymbol{\gamma}_{k}(t_{k}).
\end{equation}
The coefficient matrices are identical to those in vector space. Finally, the local state is reformulated into the global state as:
\begin{equation}~\label{equ:remap}
    \mathbf{R}(\tau) = \mathbf{R}_{k-1}\mathrm{Exp}(\boldsymbol{\theta}_{k}(\tau)),\quad
\boldsymbol{\omega}(\tau) = \mathcal{J}_{r}(\boldsymbol{\theta}_{k}(\tau))\dot{\boldsymbol{\theta}}_{k}(\tau),
\end{equation}
where $\mathrm{Exp}:\mathbb{R}^{3} \rightarrow \mathrm{SO}(3)$ is the mapping function~\cite{sola2018micro}.

\section{Multi-LiDAR Multi-IMU State Estimator}~\label{sec:lio}
Once the GPs with several motion priors on vector space and manifold are derived, we present our resilient multi-LiDAR multi-IMU state estimator by applying them into the concrete state representation.

\subsection{State Space}
Generally, it is challenging to effectively represent the state space of a multi-sensor system. One significant feature of our proposed state estimator is its ability to define the state space flexibly based on sensor configuration. Despite the presence of multiple sensors, it is sufficient to estimate the state of a primary sensor, while the states of other sensors can be determined by multiple extrinsic matrices. These matrices are pre-calibrated and remain to be fixed throughout the optimization process. For ease of subsequent derivations, we designate the world coordinate $W$ as the starting location of the primary sensor $B$. The kinematic model is described by:
\begin{equation}
\begin{array}{c}
{^{\small W}_{\small B}\dot{\mathbf{R}}(t)} = {^{\small W}_{\small B}\mathbf{R}(t)}{_{\small B}\boldsymbol{\omega}(t)}^{\wedge}, \\
{_{\small W}\dot{\mathbf{p}}(t)} = {_{\small W}\mathbf{v}(t)}, \quad 
{_{\small W}\dot{\mathbf{v}}(t)} = {_{\small W}\mathbf{a}(t)}
\end{array}
\end{equation}
where ${^{\small W}_{\small B}\mathbf{R}(t)} \in \mathrm{SO}(3)$ represents the rotation from the body coordinate to the world coordinate. ${_{\small B}\boldsymbol{\omega}(t)} \in \mathbb{R}^{3}$ is the body-frame angular velocity, and ${_{\small W}\mathbf{p}(t)}, {_{\small W}\mathbf{v}(t)}, {_{\small W}\mathbf{a}(t)} \in \mathbb{R}^{3}$ denote position, velocity, and acceleration in the world coordinate, respectively. The systems equipped with IMUs require to estimate the gyroscope bias $\mathbf{b}_{g}(t)$ and accelerometer bias $\mathbf{b}_{a}(t)$. Without causing ambiguity, we omit the subscript for simplicity in the following.

We consider the gyroscope and accelerometer as two independent sensors in our multi-LiDAR multi-IMU system, which comprises $N_{l}$ LiDARs, $N_{g}$ gyroscopes, and $N_{a}$ accelerometers, with $N_{l}\ge 1$ and $N_{g}, N_{a}\ge0$. If $N_{g}, N_{a}=0$, the estimator is simplified into a multi-LiDAR odometry. In this case, the dimensionality of state space is not influenced by the number of LiDARs but depends on the GP prior. For instance, the simplest scenario is the Random Walk prior on poses, where $\dot{\boldsymbol{\theta}}(t)=\mathbf{w}_{\theta}(t)$ and $\dot{\mathbf{p} }(t)=\mathbf{w}_{p}(t)$, leading to a minimal state space $\left \{ {\mathbf{R}(t),\mathbf{p}(t)  } \right \}\in\mathrm{SO}(3)\times \mathbb{R}^{3} $ with 6 degrees of freedom (DOF). If the rotation is assumed to have a constant-velocity prior and translation with a constant-acceleration prior, the state variables expand to $\left \{ {\mathbf{R}(t),\boldsymbol{\omega}(t),\mathbf{p}(t),\mathbf{v}(t),\mathbf{a}(t) } \right \}\in\mathrm{SO}(3)\times \mathbb{R}^{12}  $ with 15 DOF. Incorporating IMU measurements involves with adding bias onto the state variables, governed by a Random Walk Process. Specifically, with gyroscopes, the rotation state variables become $\left \{ {\mathbf{R}(t),\omega(t)},\left \{ {\mathbf{b}_{g}^{n} } \right \}^{N_{g}}_{n=1} \right \}
\in\mathrm{SO}(3)\times \mathbb{R}^{3+3N_{g}} $ using the constant-velocity prior. Similarly, for accelerometers, the translation state variables are $\left \{ \mathbf{p}(t),\mathbf{v}(t),\mathbf{a}(t) ,\left \{ {\mathbf{b}_{a}^{n} } \right \}^{N_{a}}_{n=1} \right \}
\in \mathbb{R}^{9+3N_{a}}  $ with a constant acceleration prior.

The overall state space is a combination of one rotation part and one translation part, as listed in Table~\ref{tab:state}, yielding a total number of $C_4^2 =12$ combinations. Compared to the previous $\mathrm{SE}(3)$ representation in GP state estimation~\cite{anderson2015full}, such decoupled state form enables GP to be used in inertial-aided systems beyond the limitation of LiDAR-only odometry~\cite{wu2022picking}. Moreover, our proposed approach facilitates the computation of analytic Jacobians for efficient optimization, thereby enhancing the real-time performance of the GP-based estimator.

\begin{table}[t]
\centering
\caption{State Space with different GP prior}
\label{tab:state}
\resizebox{0.5\textwidth}{!}{
\begin{threeparttable}
\begin{tabular}{@{}lccc@{}}
\toprule
\midrule
 & GP prior\tnote{1} & SDE &State Variable  \\
 \midrule
 \multirow{3}{*}{\rotatebox{90}{Rotation}}&RW&$\dot{\boldsymbol{\theta}}(t)=\mathbf{w}_{\theta}(t)$  & $\mathbf{R}(t)$  \\
  &CV&$\ddot{\boldsymbol{\theta}}(t)=\mathbf{w}_{\dot{\theta}}(t)$  & $\mathbf{R}(t),\omega(t)$  \\
   &Gyro&$\ddot{\boldsymbol{\theta}}(t)=\mathbf{w}_{\dot{\theta}}(t)$,$\dot{\mathbf{b}}_{g}^{n} (t)=\mathbf{w}_{bg}(t)$ & $\mathbf{R}(t),\boldsymbol{\omega}(t),\left \{ {\mathbf{b}_{g}^{n}(t) } \right \}^{N_{g}}_{n=1} $  \\
\midrule
   \multirow{4}{*}{\rotatebox{90}{Translation}}&RW&$\dot{\mathbf{p} }(t)=\mathbf{w}_{p}(t)$ & $\mathbf{p}(t)$  \\
   &CV&$\dot{\mathbf{w} }(t)=\mathbf{w}_{v}(t)$ & $\mathbf{p}(t),\mathbf{v}(t)$  \\
   &CA&$\dot{\mathbf{a} }(t)=\mathbf{w}_{a}(t)$ & $\mathbf{p}(t),\mathbf{v}(t),\mathbf{a}(t)$  \\
   &Accel&$\dot{\mathbf{a} }(t)=\mathbf{w}_{a}(t)$,$\dot{\mathbf{b}}_{a}^{n} (t)=\mathbf{w}_{ba}(t)$ & $\mathbf{p}(t),\mathbf{v}(t),\mathbf{a}(t),\left \{ {\mathbf{b}_{a}^{n}(t) } \right \}^{N_{a}}_{n=1}$  \\
\midrule   
\bottomrule
\end{tabular}
\begin{tablenotes}
\item[1] RW: Random Walk Process, CV: Constant-Velocity Process, CA: Constant-Acceleration Process, Gyro: CV with $N_{g}$ gyroscopes, Accel: CA with $N_{a}$ accelerometer. 
\end{tablenotes}
\end{threeparttable}
}
\vspace{-0.2in}
\end{table}

\subsection{Self-driven Kinematics}~\label{sec:piece}
Although our state representation accommodates multiple combinations for various sensor configurations, we focus on the general case that fully utilizes data from both LiDARs and IMUs. Readers can easily deduce simplified versions for LiDAR-only, LiDAR-gyroscope, and LiDAR-accelerometer scenarios with different motion priors. For a system comprising $N_{l}$ LiDARs, $N_{g}$ gyroscopes, and $N_{a}$ accelerometers, where $N_{l}, N_{g}, N_{a} > 0$, the state variables are defined as
$$
\mathbf{x}(t) \sim \left \{ \mathbf{R}(t), \boldsymbol{\omega}(t), \mathbf{p}(t), \mathbf{v}(t), \mathbf{a}(t),
\left \{ \mathbf{b}_{g}^{n} \right \}_{n=1}^{N_{g}},
\left \{ \mathbf{b}_{a}^{n} \right \}_{n=1}^{N_{a} }
\right \},
$$
residing in the space $\mathrm{SO}(3) \times \mathbb{R}^{12+3N_{g}+3N_{a}}$.

Conventional inertial navigation systems, whether employing a Kalman Filter~\cite{sola2017quaternion} or pre-integration~\cite{forster2015manifold}, typically rely on IMU measurements for state propagation~\cite{xu2022fast, shan2020lio, chen2022direct, qin2018vins}. However, these IMU-driven estimators may falter in scenarios without IMU signal or when the system's state exceeds the IMU's measuring range~\cite{he2023point}. We argue that kinematics is an inherent property of a system, while IMU data merely represents an observation. Therefore, we propose a self-driven kinematic model based on the sparse GPs described in Sec.~\ref{sec:gp}. This technique makes our state estimator independent from IMU, which is able to fuse information from multiple IMUs.

To address the nonlinearity of rotation state, we adopt a sequence of LTI SDEs to model the kinematics across the entire temporal window $[t_{0}, t_{K})$. In this piecewise approach, the window is divided into $K$ equidistant segments $\left\{ [t_{k-1}, t_{k}) \right\}_{k=1}^{K}$, where the interval length $\Delta t = t_{k} - t_{k-1}$ is a hyperparameter based on the motion profile. Within each segment, the states of the rotation are represented locally while the translation and IMU biases are in global representation. The hybrid states $\mathbf{x}_{k}(t)$ of segment $k$ is governed by the LTI SDE 
$\dot{\mathbf{x}}_{k}(t) = \mathbf{A}(t) \mathbf{x}_{k}(t) + \mathbf{F}(t)\mathbf{w}(t)$,
{
\setlength{\arraycolsep}{0pt}
\renewcommand{\arraystretch}{1.2}
$$
\begin{array}{c}
    \mathbf{x}_{k} (t)=
    \left[\begin{array}{c}
    \boldsymbol{\theta}_{k}(t)\\
    \dot{\boldsymbol{\theta}}_{k}(t)\\
    \hdashline
    \mathbf{p}_{k}(t)\\
    \mathbf{v}_{k}(t)\\
    \mathbf{a}_{k}(t)\\
    \hdashline
    \mathbf{b}_{gk}(t)\\
    \hdashline
    \mathbf{b}_{ak}(t)\\
    \end{array}\right]
    \begin{array}{l}
    \left.\begin{array}{c} \\ \\ \end{array}\right\}\text{Constant Velocity}\\
    \left.\begin{array}{c} \\ \\ \\ \end{array}\right\}\text{Constant Acceleration}\\
    \left.\begin{array}{c} \\ \end{array}\right\}\text{Random Walk}\\
    \left.\begin{array}{c} \\ \end{array}\right\}\text{Random Walk}
    \end{array},
    \mathbf{w}(t)=\begin{bmatrix}
 \mathbf{w}_{\dot{\theta}}(t)\\
\hdashline 
 \mathbf{w}_{a}(t)\\
\hdashline
 \mathbf{w}_{bg}(t)\\
\hdashline
\mathbf{w}_{ba}(t)
\end{bmatrix},\\
\mathbf{A}(t)=\mathrm{diag} (\mathbf{A}_{\small\text{CV} },\mathbf{A}_{\small\text{CA} }, 
\mathbf{A}_{\small \text{RW}}^{\small\text{bg}},\mathbf{A}_{\small \text{RW}}^{\small\text{ba}}),\\
\mathbf{F}(t)=\mathrm{diag} (\mathbf{F}_{\small\text{CV} },\mathbf{F}_{\small\text{CA} }, 
\mathbf{F}_{\small \text{RW}}^{\small\text{bg}},\mathbf{F}_{\small \text{RW}}^{\small\text{ba}}),
\end{array}
$$
}
where $\mathbf{b}_{gk}(t) \in \mathbb{R}^{3N_{g}}$ represents the stacked vectors of $N_{g}$ gyroscopes, and $\mathbf{b}_{ak}(t) \in \mathbb{R}^{3N_{a}}$  is the stacked vectors of $N_{a}$ accelerometers. The local transformation of the rotation part follows Eq.~\ref{equ:local1} and Eq.~\ref{equ:local2}. We assume constant velocity for rotation $\ddot{\boldsymbol{\theta}}_{k}(t) = \mathbf{w}_{\dot{\theta}}(t)$, constant acceleration for translation $\dot{\mathbf{a}}_{k}(t) = \mathbf{w}_{a}(t)$, and a random walk for IMU biases $\dot{\mathbf{b}}_{gk}(t) = \mathbf{w}_{bg}(t)$ and $\dot{\mathbf{b}}_{ak}(t) = \mathbf{w}_{ba}(t)$. Thus, our representation of system state consists of four parts, in which each block of the diagonal system matrices $\mathbf{A}(t)$ and $\mathbf{F}(t)$ is listed in Table~\ref{tab:prior}. Similarly, the corresponding transition and covariance matrices are denoted as below
\begin{equation}
\begin{array}{c}
\boldsymbol{\Phi}(t,t_{k-1})=\mathrm{diag} ( \boldsymbol{\Phi}_{\small\text{CV} }, \boldsymbol{\Phi}_{\small\text{CA} }, 
 \boldsymbol{\Phi}_{\small \text{RW}}^{\small\text{bg}}, \boldsymbol{\Phi}_{\small \text{RW}}^{\small\text{ba}}),\\
\mathbf{Q}(t,t_{k-1})=\mathrm{diag} (\mathbf{Q}_{\small\text{CV} },\mathbf{Q}_{\small\text{CA} }, 
\mathbf{Q}_{\small \text{RW}}^{\small\text{bg}},\mathbf{Q}_{\small \text{RW}}^{\small\text{ba}})
\end{array}
\end{equation}
We can derive the initial state at $t_{k}$ by propagating the estimated state from $t_{k-1}$ as $\mathbf{x}_{k}(t_{k}) = \boldsymbol{\Phi}(t_{k}, t_{k-1})\mathbf{x}_{k}(t_{k-1})$. For a measurement time $\tau \in [t_{k-1}, t_{k})$, we obtain the state $\mathbf{x}_{k}(\tau)$ by Gaussian interpolation $\mathbf{x}_{k}(\tau) = \boldsymbol{\Lambda}(\tau) \mathbf{x}_{k}(t_{k-1}) + \boldsymbol{\Psi}(\tau)\mathbf{x}_{k}(t_{k})$, where
\begin{equation}
\begin{array}{c}
\boldsymbol{\Lambda}(\tau)=\mathrm{diag} ( \boldsymbol{\Lambda}_{\small\text{CV} }, \boldsymbol{\Lambda}_{\small\text{CA} }, 
\boldsymbol{\Lambda}_{\small \text{RW}}^{\small\text{bg}}, \boldsymbol{\Lambda}_{\small \text{RW}}^{\small\text{ba}}),\\
\boldsymbol{\Psi}(\tau)=\mathrm{diag} ( \boldsymbol{\Psi}_{\small\text{CV} }, \boldsymbol{\Psi}_{\small\text{CA} }, 
\boldsymbol{\Psi}_{\small \text{RW}}^{\small\text{bg}}, \boldsymbol{\Psi}_{\small \text{RW}}^{\small\text{ba}})
\end{array}
\end{equation}
with each block of matrices listed in Table~\ref{tab:gp-inte}. To obtain the final interpolated global state $\mathbf{x}(\tau)$, we remap the rotation part according to Eq.~\ref{equ:remap}.

\subsection{State Estimation}
\begin{figure}[t]
	\centering
\includegraphics[width=0.5\textwidth]{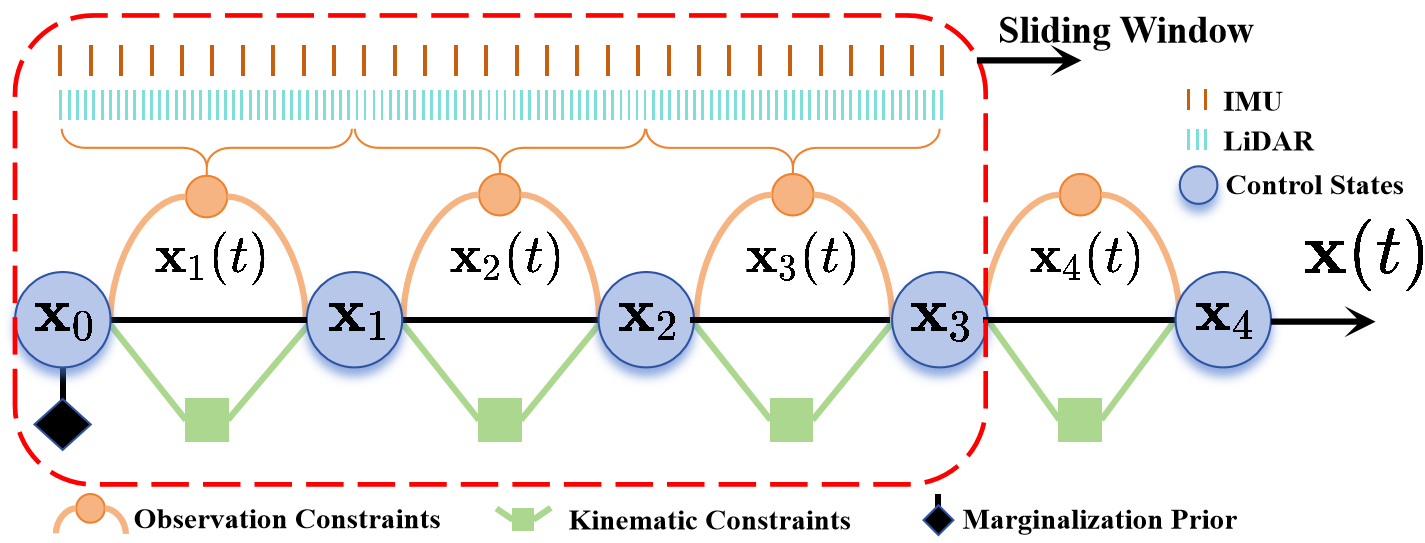}
	\caption{An illustration of sliding window optimization, including three kinds of constraints from external observation, internal kinematics, and marginalization prior. The global state $\mathbf{x}(t)$ of this sliding window consists of three segments, while the state $ \mathbf{x}_{k}(t)$ in each segment is driven by the hybrid GP prior. Upon the arrival of measurements within $[t_{3},t_{4})$, the oldest states $\mathbf{x}_{0}$ along with all measurements within $[t_{0},t_{1})$  will be marginalized.
 }
	\label{fig:sliding_window}
\end{figure}

Our multi-LiDAR with multi-IMU state estimator aims to predict the continuous-time state $\mathbf{x}(t)$ over a temporal window through a series of discrete-time measurements. Fig.~\ref{fig:sliding_window} depicts the pipeline of our continuous-time sliding-window estimator. 

We consider $M$ measurements $\mathcal{Z}=\left\{ \mathbf{z}_{i} \right\}_{i=1}^{M}$, each of which is corrupted by zero-mean Gaussian noise $\boldsymbol{\sigma}_{i} \sim (\mathbf{0}, \boldsymbol{\Sigma}_{i})$:
\begin{align}
\mathbf{z}_{i} = \mathbf{h}_{i}(\mathbf{x}(\tau_{i})) + \boldsymbol{\sigma}_{i}, \quad \tau_{i} \in [t_{0}, t_{K}).
\end{align}
where $\mathbf{h}_{i}(\cdot)$ represents the sensor-specific function, applicable to LiDAR, gyroscopes, or accelerometers. Starting with the initial mean state $ \boldsymbol{\mu}_{0}$ and prior covariance matrix $\boldsymbol{\mathcal{K}}_{0}$, our goal is to maximize the joint posterior distribution $p(\mathbf{x}(t) \mid \boldsymbol{\mu}_{0}, \mathcal{Z})$. According to Bayes' rule, this posterior can be expressed as:
\begin{equation}
    p(\mathbf{x}(t) \mid \boldsymbol{\mu}_{0}, \mathcal{Z}) \propto p(\mathcal{Z} \mid \mathbf{x}(t))p(\mathbf{x}(t) \mid \boldsymbol{\mu}_{0}).
\end{equation}

As elaborated in Sec.~\ref{sec:piece}, we utilize a piecewise LTI system to linearize the rotational state. This approach effectively manages the continuous-time states over the interval $[t_{0}, t_{K})$ using $K+1$ states $\left\{ \mathbf{x}(t_{0}), \cdots, \mathbf{x}(t_{K}) \right\}$. For the numerous measurements, we do not directly incorporate their corresponding states into the optimization process. Instead, we obtain their states $\mathbf{x}(\tau_{i})$ through GP interpolation of two consecutive control states that are predicted by the Gaussian Process Gauss-Newton optimization~\cite{tong2013gaussian}. To formalize this, we define the lift form of both the $K+1$ control states and the $M$ observations as follows:
$$
\overbrace{\begin{bmatrix}
\mathbf{x}(t_{0}) \\
\vdots \\
\mathbf{x}(t_{K})
\end{bmatrix}}^{\mathbf{x}}, \quad
\overbrace{\begin{bmatrix}
\boldsymbol{\mu}(t_{0}) \\
\vdots \\
\boldsymbol{\mu}(t_{K})
\end{bmatrix}}^{\boldsymbol{\mu}}, \quad
\overbrace{\begin{bmatrix}
\mathbf{z}_{1} \\
\vdots \\
\mathbf{z}_{M}
\end{bmatrix}}^{\mathbf{z}}, \quad
\overbrace{\begin{bmatrix}
\mathbf{h}_{1}(\mathbf{x}(\tau_{1})) \\
\vdots \\
\mathbf{h}_{M}(\mathbf{x}(\tau_{M}))
\end{bmatrix}}^{\mathbf{h}(\mathbf{x})}
$$
$$
\boldsymbol{\mathcal{K}} \doteq \left[\boldsymbol{\mathcal{K}}(t_{i}, t_{j})\right]_{i, j, 0 \le i, j \le K}, \quad
\boldsymbol{\Sigma} \doteq \mathrm{diag}(\boldsymbol{\Sigma}_{1}, \dots, \boldsymbol{\Sigma}_{M}).
$$
Maximizing a posterior of $p(\mathbf{x}(t) \mid \boldsymbol{\mu}_{0}, \mathcal{Z})$ can be formulated into a nonlinear least-square optimization problem as follows:
\begin{equation}~\label{equ:lift_opt}
\mathbf{x}^{\ast} = \arg \min_{\mathbf{x}}
\frac{1}{2}\left\|\mathbf{x} - \boldsymbol{\mu}\right\|_{\boldsymbol{\mathcal{K}}}^{2} +
\frac{1}{2}\left\|\mathbf{h}(\mathbf{x}) - \mathbf{z}\right\|_{\boldsymbol{\Sigma}}^{2}
\end{equation}
The above two terms represent the inherent kinematic prior of the system and the external observations, respectively.

\subsubsection{Inherent Kinematic Constraint}
The first term encodes the kinematics of our system, which is used to be driven by IMU information. In contrast, we drive our system by GP prior that is the foundation of our proposed versatile and resilient state estimator. Due to the Markov property of our selected priors, the inverse of kernel matrix $\boldsymbol{\mathcal{K}}^{-1}$ is exactly sparse~\cite{barfoot2014batch}. Such property allows us to simplify the error term of generic GP prior as below
\begin{equation}
    \begin{array}{c}
E_{gp}=\frac{1}{2}\left \| \mathbf{x}-\boldsymbol{\mu}  \right \|_{\boldsymbol{\mathcal{K}} }^{2}
=\frac{1}{2}\sum_{k=0}^{K} \mathbf{e}_{k}^{\top }\mathbf{Q}_{k}^{-1}\mathbf{e}_{k} ,\\
\mathbf{e}_{k}=\mathbf{x}_{k}(t_{k})-\boldsymbol{\Phi}(t_{k},t_{k-1})\mathbf{x}_{k}(t_{k-1}),
\end{array}
\end{equation}
where $\mathbf{Q}_{k}=\mathbf{Q}(t_{k},t_{k-1})$. For $k=0$, the error term is related to the prior, with $\mathbf{e}_{0}=\mathbf{x}_{0}(t_{0})-\boldsymbol{\mu}_{0}$ and $\mathbf{Q}_{0}=\boldsymbol{\mathcal{K}}_{0}$.

\subsubsection{External Observation Constraint}
The second term is related to the specific observation. Our system deals with three types of sensors, including LiDAR, gyroscope, and accelerometer. To align with our temporal segmentation, we reorganize all measurements $\mathcal{Z}$ into $K$ sets, denoted as $\mathcal{Z} \doteq \left \{ \mathcal{L}_{k},\mathcal{G}_{k},\mathcal{A}_{k}  \right \}_{k=1}^{K} $. Each set is made of measurements collected within the interval $[t_{k-1},t_{k})$ from $N_{l}$ LiDARs, $N_{g}$ gyroscopes and $N_{a}$ accelerometer. The specific forms are derived as 
$$
\begin{array}{c}
\mathcal{L}_{k}\doteq\left \{ \left\{_{k}\tilde{\mathbf{p}}_{i}^{j}, {_{k}\tau_{i}^{j}}  \right \}_{i=1}^{L_{k}^{j}}  \right \}_{j=1}^{N_{l}},\\
\mathcal{G}_{k}\doteq\left \{ \left\{_{k}\tilde{\boldsymbol{\omega}}_{i}^{j}, {_{k}\tau_{i}^{j}}\right \}_{i=1}^{G_{k}^{j}}  \right \}_{j=1}^{N_{g}},
\mathcal{A}_{k}\doteq\left \{ \left\{_{k}\tilde{\mathbf{a}}_{i}^{j}, {_{k}\tau_{i}^{j}}\right \}_{i=1}^{A_{k}^{j}}  \right \}_{j=1}^{N_{a}}, 
\end{array}
$$
where $L_{k}^{j}$ denotes the number of points measured by the $j$-th LiDAR in $k$-th segment, and  similarly for $G_{k}^{j},A_{k}^{j}$. Each measurement is a pair, consisting of the sensor's raw measurement and its corresponding timestamp ${_{k}\tau_{i}^{j}} $. By making use of the timestamp, we can precisely query the primary sensor's state $\mathbf{x}(_{k}\tau_{i}^{j})$ at $_{k}\tau_{i}^{j}$ via GP interpolation. Given the primary sensor's pose $\mathbf{T}(_{k}\tau_{i}^{j}) = \begin{bmatrix}
\mathbf{R}(_{k}\tau_{i}^{j}) & \mathbf{p}(_{k}\tau_{i}^{j}) \\
\mathbf{0} & \mathbf{1}
\end{bmatrix}$, the states of other sensors can be determined through multiplying their corresponding pre-calibrated extrinsic matrices $\mathbf{T}_{L_{j}}^{B}$, $\mathbf{T}_{G_{j}}^{B}$, and $\mathbf{T}_{A_{j}}^{B}$. In the following, we discuss the measurement model of each sensor used in our multi-LiDAR multi-IMU state estimator.

\textbf{LiDAR}: The state-of-the-art LIOs~\cite{shan2020lio,xu2022fast,chen2022direct} using the discrete-time trajectories often assume that all points are measured simultaneously at a specific timestamp. Thus, the accuracy of these systems is highly dependent on the quality of motion compensation. In contrast, our continuous-time state representation through GP allows to query the reference LiDAR pose for each point, which obviates the need for motion compensation. To ensure the effectiveness of our approach across various kinds of LiDAR, we employ a direct point-to-plane criterion for registration without tedious feature selection. The LiDAR measurement model $\mathbf{h}_{l}(\cdot)$ is defined as follows    
\begin{equation}
     \mathbf{h}_{l}(\mathbf{x}({_{k}\tau_{i}^{j}}))={_{k}\mathbf{n}_{i}^{j}}^{\top } (\mathbf{T}({_{k}\tau_{i}^{j}}) \cdot \mathbf{T}^{B}_{L_{j}} \cdot {_{k}\tilde{\mathbf{p}} _{i}^{j}}-{_{k}\mathbf{q}_{i}^{j}}  )+\boldsymbol{\sigma}_{l}.
\end{equation}
where $\boldsymbol{\sigma}_{l} \sim \mathcal{N}(\mathbf{0}, \boldsymbol{\Sigma}_{l})$ represents the additive noise, and $_{k}\mathbf{q}_{i}^{j}$ is the nearest neighbor of ${_{k}\tilde{\mathbf{p}} _{i}^{j}}$ in the map $\mathcal{M}$. The normal vector $_{k}\mathbf{n}_{i}^{j}$ is computed by Principal Component Analysis through selecting the five closest points in $\mathcal{M}$, as suggested in~\cite{xu2022fast, dellenbach2022ct}. We further discuss map management in Sec.~\ref{sec:map}. Ideally, the point-to-plane observation should approximate zero. By defining the error term of the LiDAR measurement as $\mathbf{e}_{{_{k}\tilde{\mathbf{p}}_{i}^{j}}} = \mathbf{h}_{l}(\mathbf{x}(_{k}t_{i}^{j})) - \mathbf{0}$, the energy function for the LiDAR becomes
\begin{equation}
    E_{l}  = \frac{1}{2} {\sum_{k  = 1}^{K}} \sum_{j  = 1}^{N_{l}}\sum_{i  = 1}^{L_{k}^{j}} \mathbf{e}_{{_{k}\tilde{\mathbf{p}}_{i}^{j}}}^{\top}\boldsymbol{\Sigma}_{l}^{-1}\mathbf{e}_{{_{k}\tilde{\mathbf{p}}_{i}^{j}}}
\end{equation}

\textbf{IMU}: In our self-driven system, IMU information is not required in state propagation. Instead, the IMU is treated as an external observation that directly measures the system's state. Thus, we can make use of the information from multiple IMUs. With the state at $_{k}\tau_{i}^{j}$ determined by GP interpolation, the measurement model of gyroscope is described as below:
\begin{equation}
    \mathbf{h}_{g}(\mathbf{x}(_{k}\tau_{i}^{j})) = \boldsymbol{\omega}(_{k}\tau_{i}^{j}) + \mathbf{b}_{g}^{j}(_{k}\tau_{i}^{j}) + \boldsymbol{\sigma}_{g},
\end{equation}
where $\boldsymbol{\sigma}_{g} \sim \mathcal{N}(\mathbf{0}, \boldsymbol{\Sigma}_{g})$ represents the measurement noise of gyroscope. The angular velocity $\boldsymbol{\omega}(\tau)$ of body-frame is calculated from local variables: $\boldsymbol{\omega}(\tau) = \mathcal{J}_{r}(\boldsymbol{\theta}_{k}(_{k}\tau_{i}^{j}))\dot{\boldsymbol{\theta}}_{k}(_{k}\tau_{i}^{j})$. Similarly, the measurement model of accelerometer is
\begin{equation}
     \mathbf{h}_{a}(\mathbf{x}(_{k}\tau_{i}^{j})) = \mathbf{R}(_{k}t_{i}^{j})^{-1}(\mathbf{a}(_{k}\tau_{i}^{j}) + \mathbf{g}) + \mathbf{b}_{a}^{j}(_{k}\tau_{i}^{j}) + \boldsymbol{\sigma}_{a},
\end{equation}
where $\mathbf{g}$ is the gravity in world coordinates, and $\boldsymbol{\sigma}_{a} \sim \mathcal{N}(\mathbf{0}, \boldsymbol{\Sigma}_{a})$ is the measurement noise of accelerometer. The error terms for the gyroscope and accelerometer measurements can be defined as $\mathbf{e}_{{_{k}\tilde{\omega}_{i}^{j}}} = \mathbf{h}_{g}(\mathbf{x}(_{k}\tau_{i}^{j})) - \mathbf{R}_{G_{j}}^{B}{_{k}\tilde{\boldsymbol{\omega}}_{i}^{j}}$ and $\mathbf{e}_{{_{k}\tilde{\mathbf{a}}_{i}^{j}}} = \mathbf{h}_{a}(\mathbf{x}(_{k}\tau_{i}^{j})) - \mathbf{R}_{A_{j}}^{B}{_{k}\tilde{\mathbf{a}}_{i}^{j}}$, where $\mathbf{R}_{G_{j}}^{B}$ and $\mathbf{R}_{A_{j}}^{B}$ are the rotation matrices that transfer the measurement to the coordinate of primary sensor. Consequently, the energy functions for gyroscope and accelerometer observations over the entire temporal window are formulated as below
\begin{align}
E_{g} & = \frac{1}{2} {\sum_{k  = 1}^{K}} \sum_{j  = 1}^{N_{g}}\sum_{i  = 1}^{G_{k}^{j}} \mathbf{e}_{{_{k}\tilde{\omega}_{i}^{j}}}^{\top}\boldsymbol{\Sigma}_{g}^{-1}\mathbf{e}_{{_{k}\tilde{\omega}_{i}^{j}}},\\ 
E_{a} & = \frac{1}{2} {\sum_{k  = 1}^{K}} \sum_{j  = 1}^{N_{a}}\sum_{i  = 1}^{A_{k}^{j}} \mathbf{e}_{{_{k}\tilde{\mathbf{a}}_{i}^{j}}}^{\top}\boldsymbol{\Sigma}_{a}^{-1}\mathbf{e}_{{_{k}\tilde{\mathbf{a}}_{i}^{j}}}.
\end{align}

\subsubsection{Sliding Window Optimization}
We have outlined the specific form of each error term  in Eq.~\ref{equ:lift_opt}, enabling us to reformulate the minimization as follows
\begin{equation}
    \mathbf{x}^{\ast} = \arg \min_{\mathbf{x}} \underbrace{E_{gp}}_{\text{kinematics}} + \underbrace{E_{l} + E_{g} + E_{a}}_{\text{observation}}.
\end{equation}
Typically, the above nonlinear least square minimization is solved by the iterative Gauss-Newton optimization. By the first-order Taylor expansion, a general error term around the linearization point $\bar{\mathbf{x}}$ is approximated as $\mathbf{e}(\bar{\mathbf{x}} + \delta \mathbf{x}) \approx \mathbf{e}(\bar{\mathbf{x}}) + \frac{\partial \mathbf{e}}{\partial \bar{\mathbf{x}}} \delta \mathbf{x}$, where $\delta \mathbf{x} \in \mathbb{R}^{15+3(N_{g}+N_{a})}$ is the stacked increment vector for the $K+1$ states. The Jacobian $\frac{\partial \mathbf{e}}{\partial \bar{\mathbf{x}}}$ is provided for each error term in the supplement. The optimal increment is determined by solving the normal equation $\mathbf{H}\delta \mathbf{x} = -\mathbf{b}$, where the Hessian matrix $\mathbf{H} = \sum \left( \frac{\partial \mathbf{e}}{\partial \bar{\mathbf{x}}} \right)^{\top} \mathbf{W}^{-1} \frac{\partial \mathbf{e}}{\partial \bar{\mathbf{x}}}$ and $\mathbf{b} = \sum \left( \frac{\partial \mathbf{e}}{\partial \bar{\mathbf{x}}} \right)^{\top} \mathbf{W}^{-1} \mathbf{e}$. $\mathbf{W}$ denotes the related covariance. The stack control state within the temporal window is iteratively updated as $\mathbf{x} \leftarrow \mathbf{x} + \delta \mathbf{x}$ until convergence. Notably, the rotation states are updated on the tangent space of $\mathrm{SO}(3)$, while other variables are updated directly in vector space.

To ensure real-time performance, we maintain $K$ segments and make use of a sliding window strategy. Once new measurements within the segment $[t_{k}, t_{k+1})$ are received, we shift out information from the oldest segment $[t_{0}, t_{1})$. Instead of discarding measurements from $[t_{0}, t_{1})$, we take advantage of marginalization priors to retain information beyond the current temporal window. The computation of new marginalization priors takes into account only the energy terms dependent on the marginalized variables, specifically, the measurements within $[t_{0}, t_{1})$ and old marginalization priors on state $\mathbf{x}(t_{0})$. For the detailed procedures, please refer to~\cite{demmel2021square}. More importantly, the inverse of marginalization prior matrix is obtained through the Schur complement, which is served as the exact prior covariance matrix of GP.
   
\subsection{Map Management}~\label{sec:map}
By taking advantage of the precise LiDAR measurements, the map points remain to be fixed during optimization. Typically, the system is assumed to start from a stationary state so that the map is initialized with points collected within the first 0.3 seconds. Map updating occurs in tandem with the sliding window marginalization. Once each optimization cycle converges, we update the map with points from the marginalized segment $[t_{0}, t_{1})$. To manage memory efficiently, we discard those dpoints located more than 100 meters from the current center of the primary sensor.

Various data structures have been employed to represent the map, including KD-trees~\cite{chen2022direct, xu2022fast}, octrees~\cite{hornung2013octomap}, and VDB~\cite{vizzo2022vdbfusion}. In our implementation, we adopt a spatial hashing structure as suggested in~\cite{dellenbach2022ct, vizzo2023kiss}, which offers simplicity and facilitates parallelization. The voxel size of the map is chosen with respect to the different environment. Each voxel stores a maximum number of 20 points, and the nearest-neighbor search explores the closest 7 voxels for efficiency.

\section{Experiment}
In this paper, we aim to develop a versatile and resilient state estimator tailored for multi-sensor systems, named 'Traj-LIO'. To demonstrate the versatility of Traj-LIO, we conduct comparative analysis against several state-of-the-art LIO systems using public multi-LiDAR multi-IMU dataset. To investigate the resilience of our self-driven estimator, we perform evaluations alongside other methods in scenarios involving with three distinct cases of sensor failure. Additionally, we assess the runtime performance of our Gaussian Process-based approach within the LiDAR-inertial system, highlighting its capability of real-time operation. 

The state estimator consists of three segments in the subsequent experiments, each spanning an interval of 0.04~s. Specifically, the interval is set to 0.01~s in the extreme sequence where the kinematic state surpasses the measuring range of the IMU. All evaluations are conducted on a Laptop with Intel Core i7-9750H CPU at 2.6~GHz.

\subsection{Versatility For Different Sensor Configuration}

Our LiDAR-inertial state estimator is designed for various combinations of LiDAR and IMU with different LiDAR scanning patterns. To demonstrate its effectiveness in diverse sensor configurations, we perform evaluation on the \textbf{Hilti 2021 dataset}~\cite{helmberger2022hilti}, collected by a handheld platform. The dataset provides millimeter-accurate ground truth in various settings, including indoor environments as well as outdoor scenarios. It features a 10 Hz Ouster OS0-64 LiDAR with a $360^\circ$ field of view (FoV) and a 10 Hz Livox MID70 LiDAR with a non-repetitive scan pattern within a $70^\circ$ circular FoV. In addition to the internal 100 Hz IMU in the Ouster LiDAR, the platform includes an extra 200 Hz Bosch IMU and an 800 Hz ADIS16445 IMU. We list the absolute trajectory error (ATE) results against ground truth in Table~\ref{tab:hilit}. These results are computed through a package provided by the dataset creators\footnote{https://github.com/Hilti-Research/hilti-slam-challenge-2021}.


\begin{table}[t]
\centering
\caption{{\small ATE (m) on the Hilti SLAM Challenge Dataset}}
\resizebox{0.5\textwidth}{!}{
\begin{threeparttable}
\begin{tabular}{@{}lccccccc@{}}
\toprule
\midrule
 Approach& Sensor\tnote{1}& RPG 
  & Base1
 & Base4 
 & Cons2 & Camp2  \\
\midrule
  \multirow{4}{*}{FAST-LIO} &L1,I1&0.177&0.312&0.052&0.111&0.083\\
  & L1,I2& 0.183&0.036 &\underline{0.042}  & \underline{0.063} & 0.094\\
    & L2,I1& 0.292&0.314&0.763&0.242&0.059\\
  & L2,I2& 0.267 &  0.106 &1.164  & 0.221 & 0.071\\
 \cmidrule{2-8}
   \multirow{4}{*}{Point-LIO} &L1,I1& 0.168&0.301&0.076&0.070&0.048\\
  & L1,I2& 0.176 &  \underline{0.031} &0.082  & 0.070 & 0.045\\
    & L2,I1& 5.329 &  $\times$ &0.614 & 0.230 & 0.306\\
  & L2,I2& 6.632 &  $\times$ &0.486  & 2.657 & 0.681\\
   \cmidrule{2-8}
   \multirow{4}{*}{\bf{Ours}} &L1,I1&\underline{0.167}&0.294&0.060&0.074&0.053\\
    &L1,I2&0.172&0.053&0.068&\underline{0.063}&\underline{\textbf{0.038}}\\
    &L2,I1&0.282&0.308&0.466&0.181&0.052\\
    &L2,I2&0.254&0.085&0.277& 0.166&0.058 \\
  \midrule
 \midrule
  MA-LIO&L1,L2,I2
  & 0.177&-&0.036&0.063&0.046\\
   \cmidrule{2-8}
  \multirow{2}{*}{\bf{Ours}} &L1,L2,I1& \underline{\textbf{0.163}}&0.303&\underline{\textbf{0.021}}&0.071&0.052\\
    &L1,L2,I2& 0.174&\underline{\textbf{0.030}}&0.025&\underline{\textbf{0.057}}&\underline{0.039}\\
    \midrule
    \midrule
    \multirow{3}{*}{\bf{Ours}}&L1& 0.168&0.299&0.064&0.071&0.053\\
    &L2&0.227&$\times$&0.550&$\times$&0.063\\
&L1,L2,I1,I2& 0.173&0.037&0.022&0.057&0.050\\
\midrule
\bottomrule
\end{tabular}

\begin{tablenotes}
\item[1] L1: OS0-64, L2: MID70, I1: 100~Hz IMU, I2: 200~Hz IMU.
\item[2] The best overall results are in \textbf{blod}, while \underline{underlined} is the best in each category. '$\times$' denotes divergence, and '-' represents invalid results.
\end{tablenotes}

\end{threeparttable}
}
\label{tab:hilit}
\vspace{-0.1in}
\end{table}

\subsubsection{Single LiDAR and Single IMU}

The most common configuration in LiDAR-inertial systems is the combination of a single LiDAR with a single IMU, which has been thoroughly investigated. Leading methods in this domain include FAST-LIO~\cite{xu2022fast}, Point-LIO~\cite{he2023point}, and LIO-SAM~\cite{shan2020lio}. Note that the original implementation of LIO-SAM requires a 9-axis IMU, which is incompatible with the three 6-axis IMUs in the Hilti dataset. In the case of the 64-line Ouster LiDAR, our method demonstrates the best overall performance with centimeter-level accuracy, while FAST-LIO and Point-LIO match the close performance.

Our approach particularly excels with the MID70 LiDAR. As highlighted in~\cite{zheng2023ectlo}, the MID70's non-repetitive scanning pattern and limited field of view result in more severe motion distortions and sparser point clouds compared to traditional multi-line spinning LiDARs. While FAST-LIO addresses motion distortions before registering through backpropagation, our method employs a continuous-time registration scheme. Such method allows for the iterative adjustments of point positions during optimization, leading to the promising results. In contrast, pointwise registration technique of Point-LIO appears less reliable for such sparse scanning patterns.

\subsubsection{Multi-LiDAR and Single IMU}
Recent developments in multi-LiDAR single-IMU LIO systems, such as MA-LIO~\cite{jung2023asynchronous}, SLICT~\cite{nguyen2023slict}, and CLIC~\cite{lv2023continuous}, greatly overcome the limitations of small vertical FoV in LiDAR sensors. Unfortunately, the open-source implementation of SLICT and CLIC do not provide the routines for a multi-line spinning LiDAR and a non-repetitive LiDAR. We directly report the result of MA-LIO published in~\cite{jung2023asynchronous} to facilitate a fair comparison. It can be seen that our method demonstrates the promising performance. Besides, the multiple LiDARs indeed enhances the accuracy of system in contrast to the single LiDAR setting.

\subsubsection{Other Sensor Configuration} 
Besides the configuration with a single IMU, our GP-based estimator is able to handle the input from multiple IMUs or operate with LiDAR-only setting. As shown in Table~\ref{tab:hilit}, the odoemtry performance may not increase by integrating two IMUs. This indicates that the introduction of multiple IMUs primarily enhances the robustness of LiDAR inertial odometry while the accuracy is mainly dependent on LiDAR measurements. The additional IMUs improve the resilience of the system in case of sensor failure. Furthermore, we find that our proposed LiDAR-only approach achieves the comparable results against the inertial-aided methods. This reflects that our GP prior is effective to model the self-driven kinematics.

\subsection{Resilience For Sensor Failure}
In ordinary scenarios, current LIO approaches tend to obtain the similar accuracy after fine-tuning their parameters. In real-world applications, the performance of a LiDAR-inertial state estimator lies in its resilience to inevitable sensor malfunctions. Therefore, we have conducted an in-depth investigation across three distinct types of sensor failure.

\subsubsection{No IMU Signal}
\begin{figure}[t]
	\centering
\includegraphics[width=0.45\textwidth]{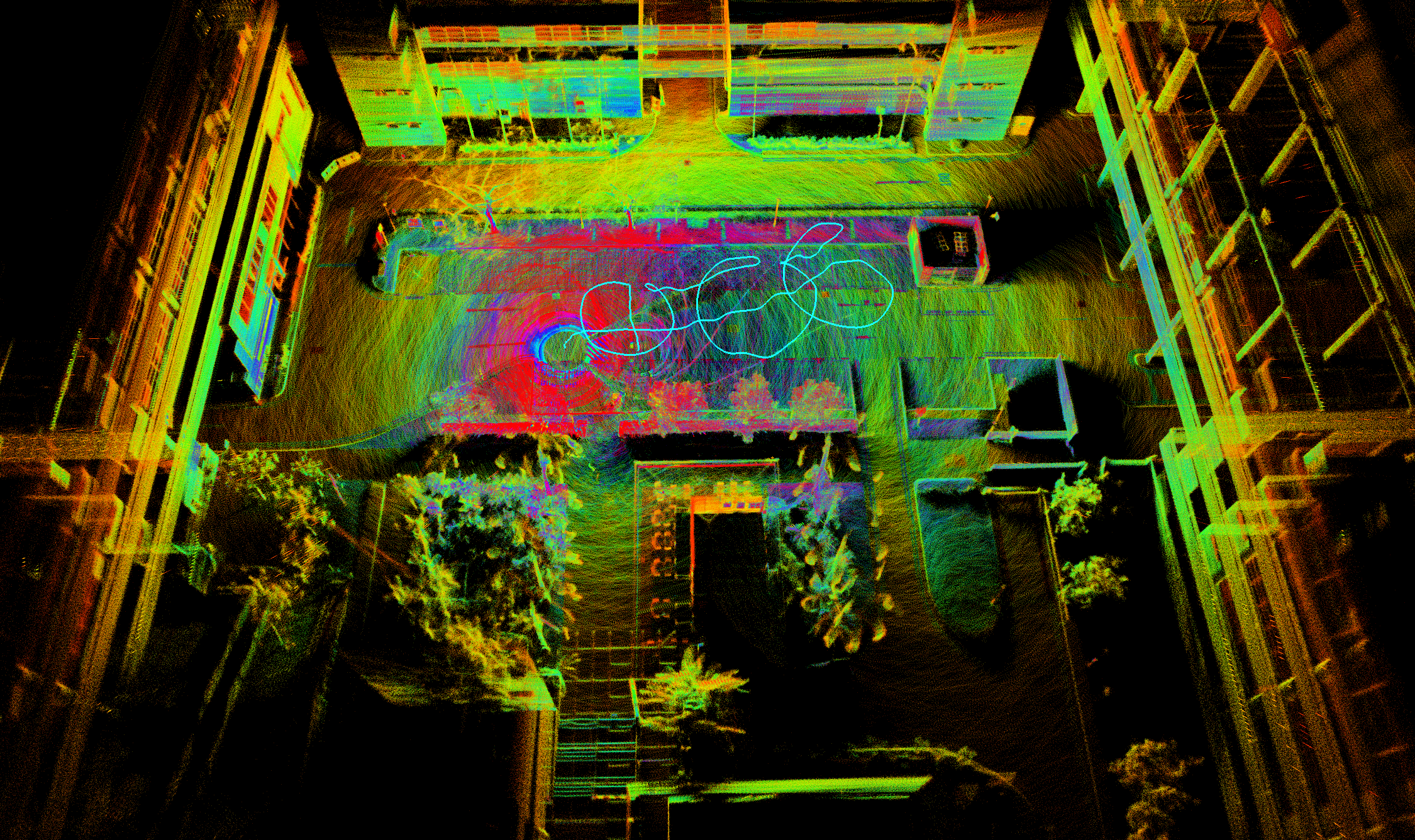}
	\caption{The mapping result of our proposed Traj-LIO on sequence eee 03 in NTU VIRAL dataset. }
	\label{fig:eee03}
	\vspace{-0.1in}
\end{figure}

\begin{table}[t]
\centering
\caption{ {\small ATE (m) on the NTU sequence eee~03 with different sensor condition}}
\label{tab:loss-signal}
\begin{threeparttable}
\begin{tabular}{@{}ccccc@{}}
\toprule
\midrule
 & LIO-SAM &FAST-LIO & Point-LIO& Ours  \\
 \midrule
 Normal&0.077&0.023&0.037&0.027\\
 Loss Gyro&$\times$\tnote{1}&$\times$&0.047&0.027\\
 Loss Accel&$\times$&$\times$&0.051&0.031 \\
 Loss IMU&$\times$&$\times$&0.053&0.030\\
\midrule
\bottomrule
\end{tabular}
\begin{tablenotes}
\item[1] $\times$ denotes failed in this condition. 
\end{tablenotes}

\end{threeparttable}
\vspace{-0.2in}
\end{table}

The widespread use of MEMS IMUs in LiDAR-inertial sensor suites is often susceptible to vibrations and temperature fluctuations, leading to potential signal delays or missing during aggressive motion profiles. To this end, we employ the \textbf{NTU VIRAL} dataset~\cite{nguyen2022ntu} to assess the odometry performance under such conditions, which is recorded on an autonomous aerial vehicle equipped with a 10 Hz horizontal Ouster OS1-16 LiDAR and a 385 Hz 9-axis IMU. To simulate the scenarios with sensor failure, we deliberately remove the input inertial information, creating cases of missing gyroscope, accelerometer, and whole IMU. To minimize the impact of no IMU signal, the evaluation is conducted on the eee~03 sequence, which captures a structural outdoor environment providing the sufficient geometric constraints for point registration, as shown in Fig.~\ref{fig:eee03}.

In Table~\ref{tab:loss-signal}, we present the ATE of different LIOs under these conditions. The first row indicates the performance under normal sensor operation, where even the least accurate system, LIO-SAM, only deviates by 0.077 meters. However, in case of sensor failure, fully IMU-driven methods like LIO-SAM and FAST-LIO fail completely. By incorporating acceleration and angular velocity into the state space, our method along with Point-LIO remain functional without IMU data. Notably, our method demonstrates better resilience to IMU failure compared to Point-LIO, maintaining similar accuracy using only LiDAR information in the structural environment.

\subsubsection{Exceeds IMU Measurement Range}
\begin{figure}[t]
    \centering
    \includegraphics[width=0.5\textwidth]{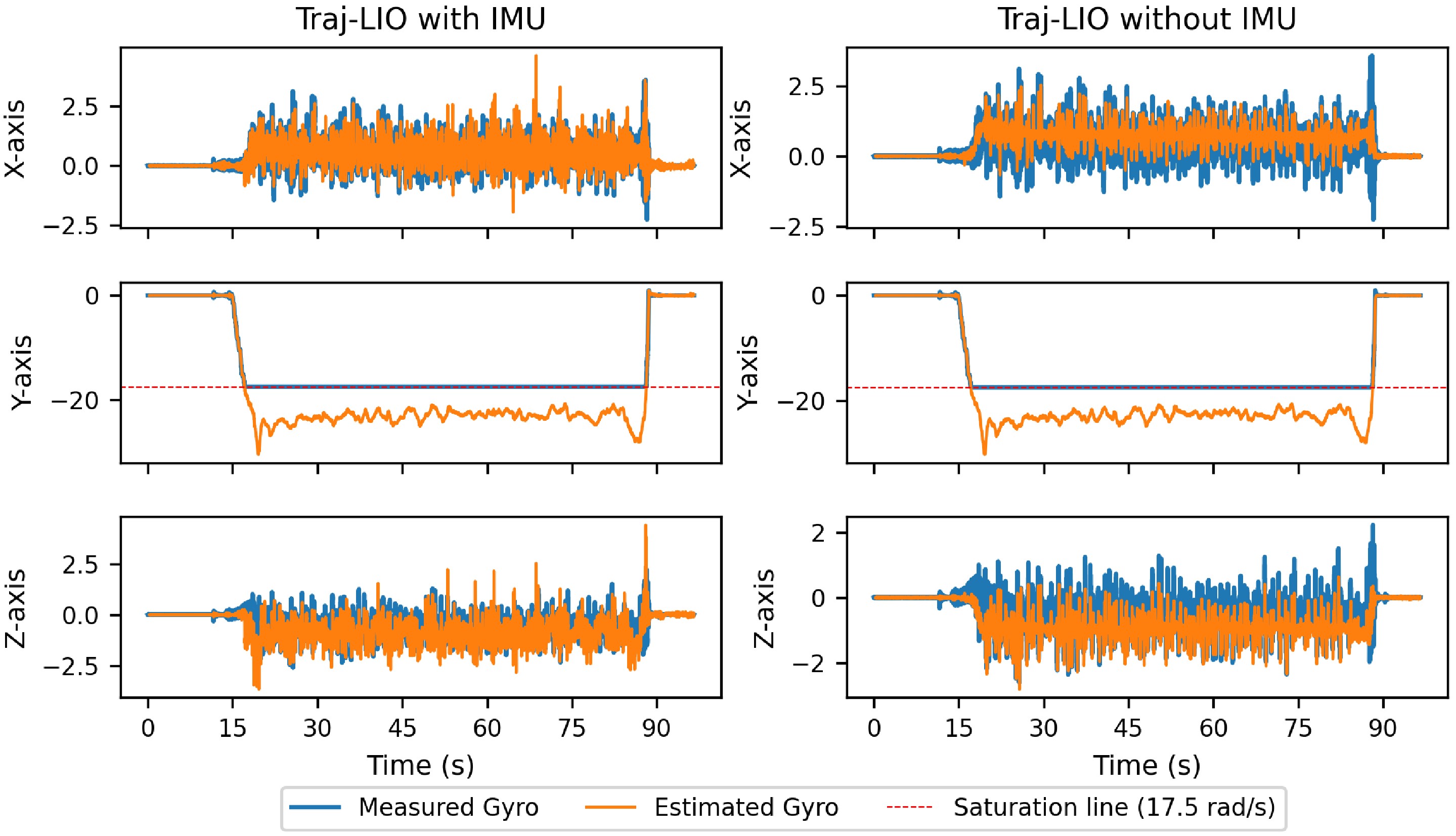}
    \caption{\small The estimated angular velocity is compared with the measured gyroscope values across three axes. The right side shows the results obtained by our method using IMU information, while the left only uses LiDAR information.
    }
    \label{fig:gyro}
\end{figure}

In addition to the aforementioned signal lost scenarios, there is another challenging situation where the IMU operates normally, yet the system's kinematic state surpasses the IMU's measurement range. To this end, we chose a sequence characterized by an aggressive spinning motion on a rotating platform in an outdoor environment, initially used in Point-LIO~\cite{he2023point}. This sequence is captured with a non-repetitive LiDAR, Livox Avia, alongside a 200 Hz IMU. It showcases prolonged periods where the gyroscope exceeds its 17.5 rad/s range, as depicted in Fig.~\ref{fig:gyro}. In such circumstances, IMU measurements fail to accurately represent the system's state, which causes the failure of IMU-driven methods like LIO-SAM and FAST-LIO.

\begin{figure}[t]
    \centering
    \includegraphics[width=0.45\textwidth]{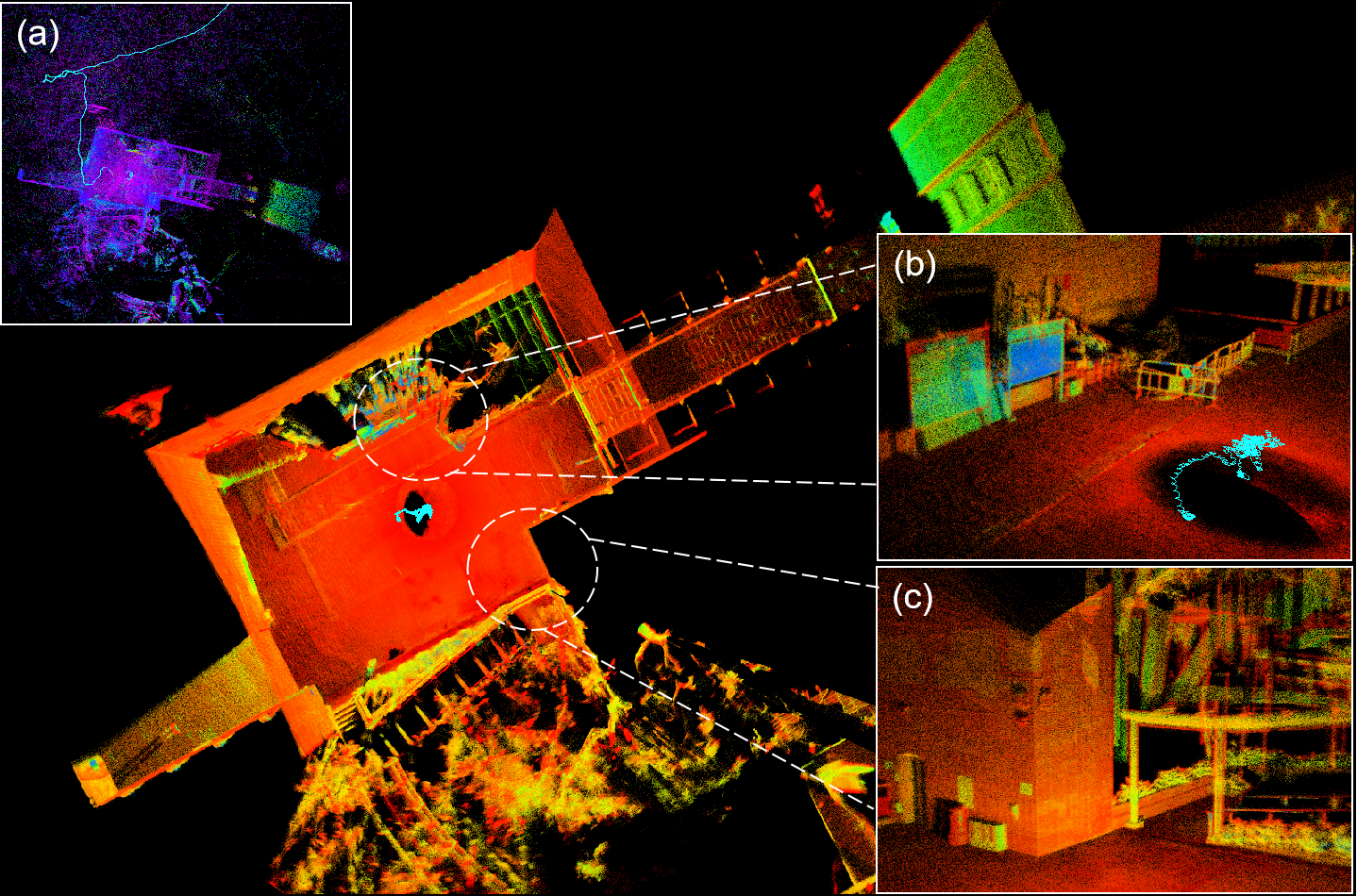}
    \caption{\small Mapping result of our method only using LiDAR information when the gyroscope exceeds its range. (a) Point-LIO fails in this scenario without IMU. (b) and (c) delve into the details of our results.}
    \label{fig:outdoor_no_gyro}
\end{figure}

By leveraging self-driven GP priors for state propagation, our presented estimator remains unaffected by erroneous IMU data. It continues to accurately predict the state of the odometry system, even when reliable gyroscope measurements are unavailable. Moreover, the GP prior effectively approximates the system's kinematics, which enables our estimator to robustly determine the system state relying solely on the geometric constraints provided by LiDAR. Fig.~\ref{fig:gyro} shows that the estimated angular velocity is nearly identical, irrespective of the availability of IMU information. In contrast, Point-LIO fails in the scenario without IMU support, as depicted in Fig.~\ref{fig:outdoor_no_gyro}.

\begin{figure*}[t]
    \centering
    \includegraphics[width=0.95\textwidth]{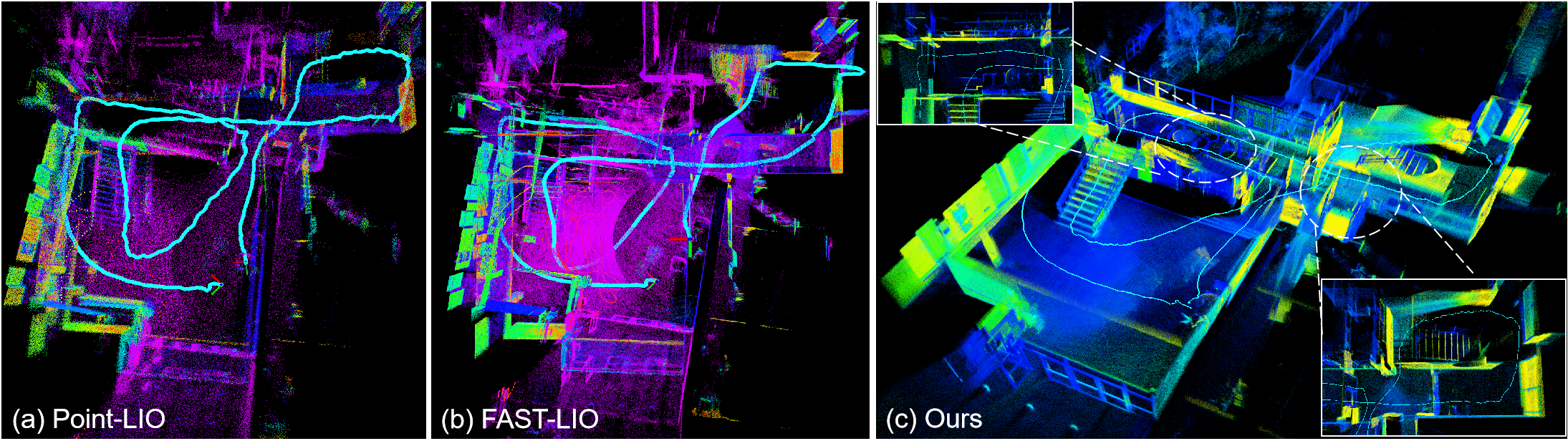}
    \caption{\small The mapping result of (a) Point-LIO, (b) FAT-LIO, and (c) our presented Traj-LIO in the multi-floor indoor scene.}
    \label{fig:indoor}
\end{figure*}

\subsubsection{LiDAR Degeneration}
Unlike the IMU failure scenarios, LiDAR data transmission remains stable through ethernet. In our assessment, the most common factor for LiDAR failure is not hardware malfunction but rather the inadequacy of the collected point cloud to provide sufficient geometric constraints for registration. To investigate this case, we collected a sequence using Livox Horizon in a multi-floor building, encompassing challenging environments such as a narrow staircase and areas with dynamic pedestrian traffic in front of the LiDAR.

We illustrate the mapping results of FAST-LIO, Point-LIO, and our Traj-LIO in Fig.~\ref{fig:indoor}. Traj-LIO demonstrates the remarkable stability and does not exhibit drift in these testing conditions by taking advantage of our  proposed multi-state constraint approach. Conversely, the filter-based methods, FAST-LIO and Point-LIO, encounter significant localization errors, which are severely affected by indoor degeneration scenarios and the presence of dynamic objects.

\subsection{Runtime Analysis}
\begin{table}[t]
\centering
\caption{\small Processing time of sequence RPG (89s) from Hilti 2021.}
\label{tab:runtime}

\resizebox{0.5\textwidth}{!}{
\begin{threeparttable}
\begin{tabular}{@{}lccccc@{}}
\toprule
\midrule
 LiDAR&Point Rate (p/s)& FAST-LIO &Point-LIO & Our-seg1\tnote{1} &Our-seg4 \\
 \midrule
 OS0-64&2,621,44&16.3&24.9&20.9&79.2\\
 Mid-70& 100,000 &4.3&11.7&16.4&39.0\\

\midrule
\bottomrule
\end{tabular}
\begin{tablenotes}
\item[1] seg1 has one segment in the optimization window while seg4 has 4. The segment interval is 0.04~s. 
\item[2] ALL methods use single LiDAR with 100~Hz IMU. 
\end{tablenotes}
\end{threeparttable}
} 
\vspace{-0.2in}
\end{table}

The heavy computational cost hinders the previous GP-based estimator from the real-time applications. To address this challenge, we develop a multi-sensor state estimator to facilitate real-time applications. Specifically, we separate the rotation and translation components in the state space, which enables us to derive analytic Jacobians on vector space and $\mathrm{SO}(3)$ in order to avoid the complexities associated with $\mathrm{SE}(3)$. We give the details on all the analytical Jacobians in our supplementary material. Moreover, we make use of parallel processing techniques like the TBB library to handle the computational load of numerous nearest neighbor searches.

To showcase real-time capabilities of our method, we selected RPG sequence from Hilti dataset with different kinds of LiDAR. As shown in Table~\ref{tab:runtime}, our total processing time consistently stays below the sequence duration of 89 s. When the segment is set to one, our processing time is on par with the discrete-time methods while the proposed continuous-time state estimator through GP offers a versatile and resilient solution for multi-sensor fusion. Admittedly, our estimator with multiple segments incurs the extra computational load, however, there still remains considerable potential for code optimization to enhance its efficiency.

\section{Conclusion and Future Work}~\label{sec:conc}
This paper has introduced a versatile and resilient multi-LiDAR multi-IMU state estimator under the Gaussian Process framework that not only offers a non-parametric continuous-time trajectory representation but also precisely models the system's inherent kinematics through GP priors. Moreover, our presented method is able to integrate multiple asynchronous sensors and exhibit robustness against inevitable sensor failures. To enhance real-time performance, we employ three different types of priors to minimally and precisely model states of rotation, translation, and IMU bias. Extensive experiments demonstrate the efficacy of our presented multi-LiDAR multi-IMU state estimator.

Although having achieved the real-time performance on LiDAR-inertial odometry, our proposed approach still suffers from the heavy computational burden with the increasing number of segments. In future work, we aim to explore the potential of sparsity within multi-state sliding window optimization and incorporate a more efficient map structure.

\bibliographystyle{plainnat}
\bibliography{references}

\end{document}